\pdfoutput=1
\documentclass[manuscript,screen]{acmart}
\usepackage{wrapfig}

\AtBeginDocument{%
  \providecommand\BibTeX{{%
    \normalfont B\kern-0.5em{\scshape i\kern-0.25em b}\kern-0.8em\TeX}}}
\setcopyright{acmcopyright}

\setcopyright{acmlicensed}
\copyrightyear{2020}
\acmYear{2020}
\acmDOI{10.1145/3383313.3412217}

\acmConference[RecSys '20]{Fourteenth ACM Conference on Recommender Systems}{September 22--26, 2020}{Virtual Event, Brazil}
\acmBooktitle{Fourteenth ACM Conference on Recommender Systems (RecSys '20), September 22--26, 2020, Virtual Event, Brazil}
\acmPrice{15.00}
\acmISBN{978-1-4503-7583-2/20/09}

\begin{document}

\title{Carousel Personalization in Music Streaming Apps with Contextual Bandits}

\author{Walid Bendada}
\author{Guillaume Salha}
\author{Th\'{e}o Bontempelli}
\affiliation{%
  \institution{Deezer}
  \city{Paris}
  \country{France}
}
\email{research@deezer.com}

\renewcommand{\shortauthors}{W. Bendada, et al.}
\newcommand{\up}[1]{\textsuperscript{#1}}

\begin{abstract}
Media services providers, such as music streaming platforms, frequently leverage swipeable carousels to recommend personalized content to their users. However, selecting the most relevant items (albums, artists, playlists...) to display in these carousels is a challenging task, as items are numerous and as users have different preferences. In this paper, we model carousel personalization as a contextual multi-armed bandit problem with multiple plays, cascade-based updates and delayed batch feedback. We empirically show the effectiveness of our framework at capturing characteristics of real-world carousels by addressing a large-scale playlist recommendation task on a global music streaming mobile app. Along with this paper, we publicly release industrial data from our experiments, as well as an open-source environment to simulate comparable carousel personalization learning problems.
\end{abstract}

\ccsdesc[300]{Information systems~Recommender systems}
\ccsdesc[300]{Information systems~Personalization}
\ccsdesc[300]{Information systems~Content ranking}
\ccsdesc[300]{Computing methodologies~Learning paradigms}
\ccsdesc[300]{Computing methodologies~Online learning settings}
\ccsdesc[300]{Computing methodologies~Sequential decision making}

\keywords{Multi-Armed Bandits with Multiple Plays, Contextual Bandits, Cascade Models, Expected Regret, Carousel Personalization, Playlist Recommendation, Music Streaming Platforms, A/B Testing}

\maketitle

\section{Introduction}
\label{s1}

Recommending relevant and personalized content to users is crucial for media services providers, such as news \cite{li2010contextual}, video \cite{covington2016deep} or music streaming \cite{schedl2018current} platforms. Indeed, effective recommender systems improve the users' experience and engagement on the platform, by helping them navigate through massive amounts of content, enjoy their favorite videos or songs, and discover new ones that they might like \cite{bobadilla2013recommender,zhang2019deep,schedl2018current}. As a consequence, significant efforts were initiated to transpose promising research on these aspects to industrial-level applications \cite{covington2016deep,gomez2015netflix,jacobson2016music,lerallut2015large,mcinerney2018explore,ying2018graph}.

In particular, many global mobile apps and websites, notably from the music streaming industry, currently leverage \textit{swipeable carousels} to display recommended content on their homepages. These carousels, also referred to as \textit{sliders} or \textit{shelves} \cite{mcinerney2018explore}, consist in ranked lists of items or \textit{cards} (albums, artists, playlists...). A few cards are initially displayed to the users, who can click on them or swipe on the screen to see some of the additional cards from the carousel. Selecting and ranking the most relevant cards to display is a challenging task \cite{ma2015introduction,mcinerney2018explore,gruson2019offline,huang2019contextual}, as the catalog size is usually significantly larger than the number of available slots in a carousel, and as users have different preferences. While being close to \textit{slate recommendation} \cite{kale2010non,ie2019slateq,swaminathan2017off,jiang2018optimizing} and to \textit{learning to rank} settings \cite{radlinski2008learning,liu2009learning,pereira2019online}, carousel personalization also requires dealing with user feedback to adaptively improve the recommended content via \textit{online learning} strategies \cite{hoi2018online,anantharam1987asymptotically,chen2016combinatorial}, and integrating that some cards from the carousel might not be seen by users due to the swipeable structure.

In this paper, we model carousel personalization as a  multi-armed bandit with multiple plays \cite{anantharam1987asymptotically} learning problem. Within our proposed framework, we account for important characteristics of real-world swipeable carousels, notably by considering that media services providers have access to contextual information on user preferences, that they might not know which cards from a carousel are actually seen by users, and that feedback data from carousels might not be available in real time. Focusing on music streaming applications, we show the effectiveness of our approach by addressing a large-scale carousel-based playlist recommendation task on the global mobile app Deezer\footnote{\href{https://www.deezer.com}{https://www.deezer.com}}. With this paper, we also release industrial data from our experiments, as well as an open-source environment to simulate comparable carousel personalization learning problems. This paper is organized as follows. In Section \ref{s2}, we introduce and formalize our multi-armed bandit framework for carousel personalization. We detail our data, our playlist recommendation task and our experimental setting in Section~\ref{s3}. We present and discuss our results in Section \ref{s4}, and we conclude in Section \ref{s5}.

\section{A Contextual Multi-Armed Bandit Framework for Carousel Personalization}
\label{s2}

In this section, after reviewing key notions on multi-armed bandits with multiple plays, we introduce our framework.

\subsection{Background on Multi-Armed Bandits with Multiple Plays}
\label{s21}

Multi-armed bandits are among the most famous instances of sequential decision making problems \cite{kuleshov2014algorithms,slivkins2019introduction,sutton2011reinforcement}. Multi-armed bandits \textit{with multiple plays} \cite{anantharam1987asymptotically,komiyama2015optimal} involve $K$ entities called \textit{arms}. At each round $t = 1, 2, . . . , T$, a \textit{forecaster} has to select \textit{a set} $S_t \subset \{1,...,K\}$ of $L < K$ arms (while $L = 1$ in the \textit{single play} version of the problem \cite{slivkins2019introduction}). The forecaster then receives some \textit{rewards} from the selected arms, that we assume to be binary. The reward associated to an arm $i \in S_t$ is a sample drawn from a Bernoulli($p_i$) distribution, with $p_i \in [0,1]$ being an unknown parameter. Bernoulli distributions of arms $1,...,K$ are assumed independent, which we later discuss. The objective of the forecaster is to maximize the sum of rewards received from the selected arms over time. It requires identifying the optimal set $\delta^*(L) \subset \{1,...,K\}$ of the $L$ arms associated to the top-$L$ highest Bernoulli parameters, i.e. the $L$ highest expected rewards, as fast as possible.

In such problems, the forecaster faces an \textit{exploration-exploitation dilemma}. As the environment does \textit{not} reveal the rewards of the unselected arms, the forecaster needs to try all arms over time to identify the best ones (\textit{exploration}). However, selecting underperforming arms also leads to lower expected rewards, which encourages the forecaster to repeatedly select the assumed best ones (\textit{exploitation}). Over the past years, several strategies have been proposed and studied, aiming at providing efficient trade-offs between these two opposite objectives when sequentially selecting sets $S_t$. Notable examples include the Upper Confidence Bound (UCB) \cite{auer2002finite,chen2016combinatorial,lai1985asymptotically,wang2018regional} and Thompson Sampling \cite{komiyama2015optimal,chapelle2011empirical,thompson1933likelihood} algorithms (see Section \ref{s3}). The \textit{expected cumulative regret} $\text{Reg}(T)= \sum_{t=1}^T \Big( \sum_{i \in \delta^*(L)} p_i - \sum_{i \in S_t} p_i \Big)$, which represents the expected total loss endured by the forecaster by selecting non-optimal sets of arms at rounds 1 to $T$, is a common measure to compare the performances of strategies addressing this top-$L$ best arms identification problem \cite{sutton2011reinforcement,slivkins2019introduction,anantharam1987asymptotically,komiyama2015optimal,chen2016combinatorial,wang2018regional}.

\subsection{Multi-Armed Bandits with Multiple Plays for Carousel Personalization}
\label{s22}

Throughout this paper, the $K$ arms will correspond to a list of $K$ cards/items, such as a catalog of albums or playlists in a music streaming app. They can be recommended to $N$ users through a swipeable carousel containing $L \ll K$ slots. As users have various preferences, different cards can be displayed to different users. The $L$ recommended cards from the carousel of each user, i.e. the $L$ \textit{selected arms} for each user, are updated at regular intervals or \textit{rounds}, whose frequency depends on the technical constraints of the platform. We aim at optimizing \textit{display-to-stream rates}, i.e. at identifying the $L$ cards for which each user is the most likely to click and then to \textit{stream} the underlying content, \textit{at least once} during the round. When a card $i$ is displayed to a user $u$, such streaming activity, i.e. a reward of 1, occurs during the round with an unknown probability $p_{ui} \in [0,1]$. Here, we assume that the number of cards, the number of users, and the display-to-stream probabilities $p_{ui}$ are fixed ; we later discuss these assumptions. A naive way to tackle this problem would consist in simultaneously running $N$ standard bandit algorithms, aiming at individually identifying the top-$L$ cards with highest $p_{ui}$ probabilities for each user $u$. This approach is actually unsuitable and would require a too long training time to reach convergence. Indeed, the number of display-to-stream parameters to estimate would be $K \times N$, which is very large in practice as platforms often have millions of active users. In Section \ref{s23}, we describe two strategies to address this problem by leveraging \textit{contextual information} on user preferences.

\subsection{Leveraging Contextual Information on User Preferences}
\label{s23}

\subsubsection{Semi-Personalization via User Clustering}
\label{s231}

First, let us assume that we have access to a \textit{clustering of users}, constructed from users' past behaviours on the platform. Each user belongs to one of the $Q$ groups $C_1, C_2,...,C_Q$ with $Q \ll N$. For instance, on a music streaming app, users from a same group would have homogeneous musical tastes. We propose to assume that users from a same group have identical expected display-to-stream probabilities for each card:
\begin{equation}
\forall c \in \{C_1,...,C_Q\}, \forall u \in c, \forall i \in \{1,...,K\}, p_{ui} = p_{ci}.
\end{equation}
Then, we simultaneously run $Q$ bandit algorithms, one for each cluster, to identify the top-$L$ best cards to recommend to each group. This strategy reduces the number of parameters to estimate to $K \times Q$, which is significantly fewer than $K \times N$ in practice. Moreover, thanks to such users gathering, platforms receive more feedback on each displayed card w.r.t. the previous naive setting. This ensures a faster and more robust identification of optimal sets. However, the empirical performance of this strategy also strongly depends on the quality of the underlying user clustering.

\subsubsection{Contextual Multi-Armed Bandits}
\label{s232}

Instead of relying on clusters, let us now assume that we directly have access to a $D$-dimensional attribute vector $x_u \in \mathbb{R}^D$ for each user $u$. These vectors aim at summarizing user preferences on the platform, e.g. their musical tastes (in terms of genres, moods, countries...) for a music streaming app. We assume that the expected display-to-stream probabilities of a user $u$ are functions of his/her attribute vector:
\begin{equation}
\forall i \in \{1,...,K\}, p_{ui} = \sigma(x_u^T \theta_i),
\end{equation}
where $\theta_1,...,\theta_K$ are $D$-dimensional weight vectors to learn for each of the $K$ arms, and where $\sigma(\cdot)$ is the sigmoid function: $\sigma(x) = 1/(1 + e^{-x})$. This corresponds to the \textit{contextual bandit} setting \cite{li2010contextual,chu2011contextual,agarwal2014taming}, a popular learning paradigm for online recommender systems \cite{li2010contextual,tang2014ensemble,zhou2016latent,mcinerney2018explore,gruson2019offline,wang2017efficient,zoghi2017online,qin2014contextual,li2011unbiased}. Strategies to learn weight vectors are detailed e.g. in \cite{chapelle2011empirical,mcinerney2018explore}. As $D \ll N$ in practice, such strategy also significantly reduces the number of parameters, to $K \times D$. By design, users with similar preferences will have close expected display-to-stream probabilities. Moreover, all $N$ users can end up with different optimal carousels, contrary to the aforementioned semi-personalized clustering approach. 

\subsection{Capturing Characteristics of Real-World Carousels: Cascade-Based Updates, Delayed Feedback}
\label{s24}

In our framework, we also aim at capturing other important characteristics of real-world swipeable carousels. In particular, while standard bandit algorithms usually consider that the forecaster receives rewards (0 or 1) from \textit{each} of the $L$ selected arms at each round, in our setting some selected cards might actually \textit{not} be seen by users. As illustrated in Figure~\ref{fig:carousel}, only a few cards, say $L_{\text{init}} < L$, are initially displayed on a user's screen. The user needs to swipe right to see additional cards. As we later verify, ignoring this important aspect, and thus returning a reward of 0 for all unclicked cards at each round whatever their rank in the carousel, would lead to underestimating display-to-stream probabilities.

In this paper, we assume that we do \textit{not} exactly know how many cards were seen by each user. Such assumption is consistent with Deezer's actual usage data and is realistic. Indeed, on many real-world mobile apps carousels, users usually do not click on any button to discover additional cards, but instead need to continuously swipe left and right on the screen. As a consequence, the card display information is ambiguous, and is technically hard to track with accuracy.

Here, to address this problem, we consider and later evaluate a \textit{cascade-based arm update} model. We draw inspiration from the \textit{cascade model} \cite{craswell2008experimental}, a popular approach to represent user behaviours when facing ranked lists of recommended items in an interface, with numerous applications and extensions \cite{kveton2015cascading,katariya2016dcm,zong2016cascading,lagree2016multiple}. At each round, we consider that:
\begin{itemize}
    \item An active user who did not stream any card during the round only saw the $L_{\text{init}}$ first ones. 
    \item An active user who streamed the i\up{th} card, with $i \in \{1,...,L\}$, saw all cards from ranks 1 to $\text{max}(L_{\text{init}},i)$.
\end{itemize}
For instance, let $L_{\text{init}}=3$ and $L=12$. The reward vectors obtained from users who a) did not stream during the round, b) only streamed the 2\up{nd} card, and c) streamed the 2\up{nd} and 6\up{th} cards, are as follows, with $X$ denoting no reward:
\begin{center}
    $a:~[0, 0, 0, X, X , X, X, X, X, X, X, X] \hspace{0.3cm}$ $b:~[0, 1, 0, X, X, X, X, X, X, X, X, X] \hspace{0.3cm}$ $c:~[0, 1, 0, 0, 0, 1, X, X, X, X, X, X]$
\end{center}

Last, to be consistent with real-world constraints, we assume that rewards are not processed on the fly but by \textit{batch}, at the end of each round e.g. every day. We study the impact of such \textit{delayed batch feedback} in our upcoming experiments.

\subsection{Related Work}
\label{s25}

Bandits are very popular models for online recommendation \cite{pereira2019online,radlinski2008learning,li2010contextual,qin2014contextual,tang2014ensemble,wang2017efficient,zhou2016latent,li2011unbiased,nguyen2014dynamic}. In particular, \cite{gruson2019offline} and  \cite{mcinerney2018explore} also recently studied carousel personalization in mobile apps. \cite{mcinerney2018explore} introduced a contextual bandit close to our Section
~\ref{s232}. However, their approach focuses more on explainability, they do not model cascade-based displays and do not integrate semi-personalized strategies. \cite{gruson2019offline} also considered contextual bandits inspired from \cite{mcinerney2018explore} for playlist recommendation in carousels, but did not provide details on their models. They instead aimed at predicting the online ranking of these models from various offline evaluations. Last, other different sets of ordered items have been studied \cite{ie2019slateq,jiang2018optimizing,kale2010non,zong2016cascading,swaminathan2017off,jiang2018beyond}.

\section{Experimental Setting}
\label{s3}

In the following, we empirically evaluate and discuss the effectiveness of our carousel personalization framework.

\subsection{Playlist Recommendation on a Global Music Streaming App}
\label{s31}

We study a large-scale carousel-based playlist recommendation task on the global mobile app Deezer. We consider $K = 862$ playlists, that were created by professional curators from Deezer with the purpose of complying with a specific music genre, cultural area or mood, and that are among the most popular ones on the service. Playlists' cover images constitute the cards that can be recommended to users on the app homepage in a carousel, updated on a daily basis, with $L = 12$ available slots and $L_{\text{init}}=3$ cards initially displayed. Figure~\ref{fig:carousel} provides an illustration of the carousel.

To determine which method would best succeed in making users click and stream the displayed playlists, extensive experiments were conducted in two steps. First, \textit{offline} experiments simulating users' responses to carousel-based recommendations were run, on a simulation environment and on data that we both publicly release\footnote{\label{code} Data and code are available at: \href{https://github.com/deezer/carousel_bandits}{https://github.com/deezer/carousel\_bandits}} with this paper (see Section \ref{s32}). We believe that such industrial data and code release  will benefit the research community and future works. Then, an \textit{online} large-scale A/B test was run on the Deezer app to validate the findings of offline experiments.

\subsection{A Simulation Environment and Dataset for Offline Evaluation of Carousel-Based Recommendation}
\label{s32}

For offline experiments, we designed a simulated environment in Python based on 974 960 fully anonymized Deezer users. We release a dataset in which each user $u$ is described by a feature vector $x_u$ of dimension $D = 97$, computed internally by factorizing the interaction matrix between users and songs as described in \cite{hu2008collaborative} and then adding a bias term. A $k$-means clustering with $Q= 100$ clusters was also performed to assign each user to a single cluster. In addition, for each user-playlist pair, we release a "ground-truth" display-to-stream probability $p_{ui} = \sigma(x_u^T \theta_i)$ where, as in \cite{chapelle2011empirical}, the $D$-dimensional vectors $\theta_i$ were estimated by fitting a logistic regression on a click data history from January 2020.

Simulations proceed as follows. At each round, a random subset of users (20 000, in the following) is presented to several sequential algorithms a.k.a. policies to be evaluated. These policies must then recommend an ordered set of $L = 12$ playlists to each user. Streams, i.e. positive binary rewards, are generated according to the aforementioned display-to-stream probabilities and to a configurable cascading browsing model capturing that users explore the carousel from left to right and might not see all recommended playlists. At the end of each round, all policies update their model based on the set of users and on binary rewards received from displayed playlists. Expected cumulative regrets of policies \cite{sutton2011reinforcement,anantharam1987asymptotically,komiyama2015optimal} w.r.t. the optimal top-$L$ playlists sets according to $p_{ui}$ probabilities are computed.

\subsection{Algorithms}
\label{s33}

\def \randomAlgo{\textit{random}}
\def \exploreThenCommitAlgo[#1]{\textit{etc-seg#1}}
\def \exploreThenCommitExplore[#1]{\textit{etc-seg-explore#1}}
\def \exploreThenCommitExploit[#1]{\textit{etc-seg-exploit#1}}
\def \epsilonGreedyAlgo[#1]{\textit{$\epsilon$-greedy-seg#1}}
\def \epsilonGreedyExplore[#1]{\textit{$\epsilon$-greedy-seg-explore#1}}
\def \epsilonGreedyExploit[#1]{\textit{$\epsilon$-greedy-seg-exploit#1}}
\def \klucbAlgo{\textit{kl-ucb-seg}}
\def \TSAlgo[#1]{\textit{ts-seg#1}}
\def \LinearTSAlgo[#1]{\textit{ts-lin#1}}

In our experiments, we evaluate semi-personalized versions of several popular sequential decision making algorithms/policies, using the provided $Q=100$ clusters, and compare their performances against fully-personalized methods. As detailed in Section \ref{s231}, users within a given cluster share parameters for all semi-personalized policies; they are the ones whose names end with \textit{-seg} in the following list. We consider the following methods:

\begin{itemize}
    \item \randomAlgo{}: a simple baseline that randomly recommends $L$ playlists to each user.
    \item \epsilonGreedyAlgo[]: recommends playlists randomly with probability $\epsilon$, otherwise recommends the top-$L$ with highest mean observed rewards. Two versions, \epsilonGreedyExplore[] ($\epsilon$=0.1) and \epsilonGreedyExploit[] ($\epsilon$=0.01) are evaluated.
    \item \exploreThenCommitAlgo[]: an \textit{explore then commit} strategy, similar to \randomAlgo{} until all arms have been played $n$ times, then recommends the top-$L$ playlists. Two versions, \exploreThenCommitExplore[] ($n=100$) and \exploreThenCommitExploit[] ($n=20$) are evaluated.
    \item \klucbAlgo{}: the Upper Confidence Bound (UCB) strategy \cite{auer2002finite,chen2016combinatorial,lai1985asymptotically}, that tackles the exploration-exploitation trade-off by computing confidence intervals for the estimation of each arm probability, then selecting the $L$ arms with highest upper confidence bounds. Here, we use KL-UCB bounds \cite{garivier2011kl}, tailored for Bernoulli rewards. 
    \item \TSAlgo[]: the Thompson Sampling strategy \cite{chapelle2011empirical,thompson1933likelihood}, in which estimated display-to-stream probabilities are samples drawn from Beta distributions \cite{thompson1933likelihood}, whose parameters are updated at each round in a Bayesian fashion, such that variance tends towards zero and expectation converges to empirical mean as more rewards are observed. Two versions, \textit{ts-seg-naive} (prior distributions are Beta$(1,1)$, i.e. Uniform$(0,1)$) and \textit{ts-seg-pessimistic} (priors are Beta$(1,99)$) are evaluated. As the UCB algorithm \cite{chen2016combinatorial}, Thompson Sampling is backed by strong theoretical guarantees \cite{komiyama2015optimal} on  speeds of expected cumulative regrets in the multi-armed bandit with multiple plays setting.
    \item \LinearTSAlgo[]: an extension of Thompson Sampling \cite{chapelle2011empirical} to the linear contextual framework from Section \ref{s232}. We follow the method of \cite{chapelle2011empirical} to learn $\theta_{i}$ vectors for each arm $i$ from Gaussian prior distributions. Two versions, \textit{ts-lin-naive} (0 means for all dimensions of the prior) and \textit{ts-lin-pessimistic} (-5 mean for the bias dimension prior) are evaluated.
\end{itemize}

By default, policies always abide by the cascade model introduced in Section \ref{s24}, meaning they do not update the parameters relative to recommended playlists that the cascade model labels as unseen. For comparison, we also implemented versions of these policies that do not abide by this behaviour. In the following, they are labelled \textit{no-cascade}.

\section{Experimental Results}
\label{s4}

\subsection{Offline Evaluation}
\label{s41}
\begin{figure*}[t]
\minipage{0.3\textwidth}
\centering
  \includegraphics[width=0.7\textwidth]{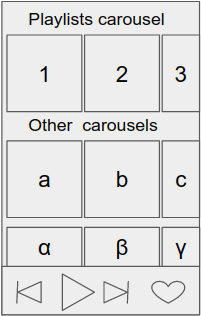}
  \caption{Swipeable carousels on the app.}\label{fig:carousel}

\endminipage\hfill
\minipage{0.68\textwidth}
\centering
  \includegraphics[width=0.98\textwidth]{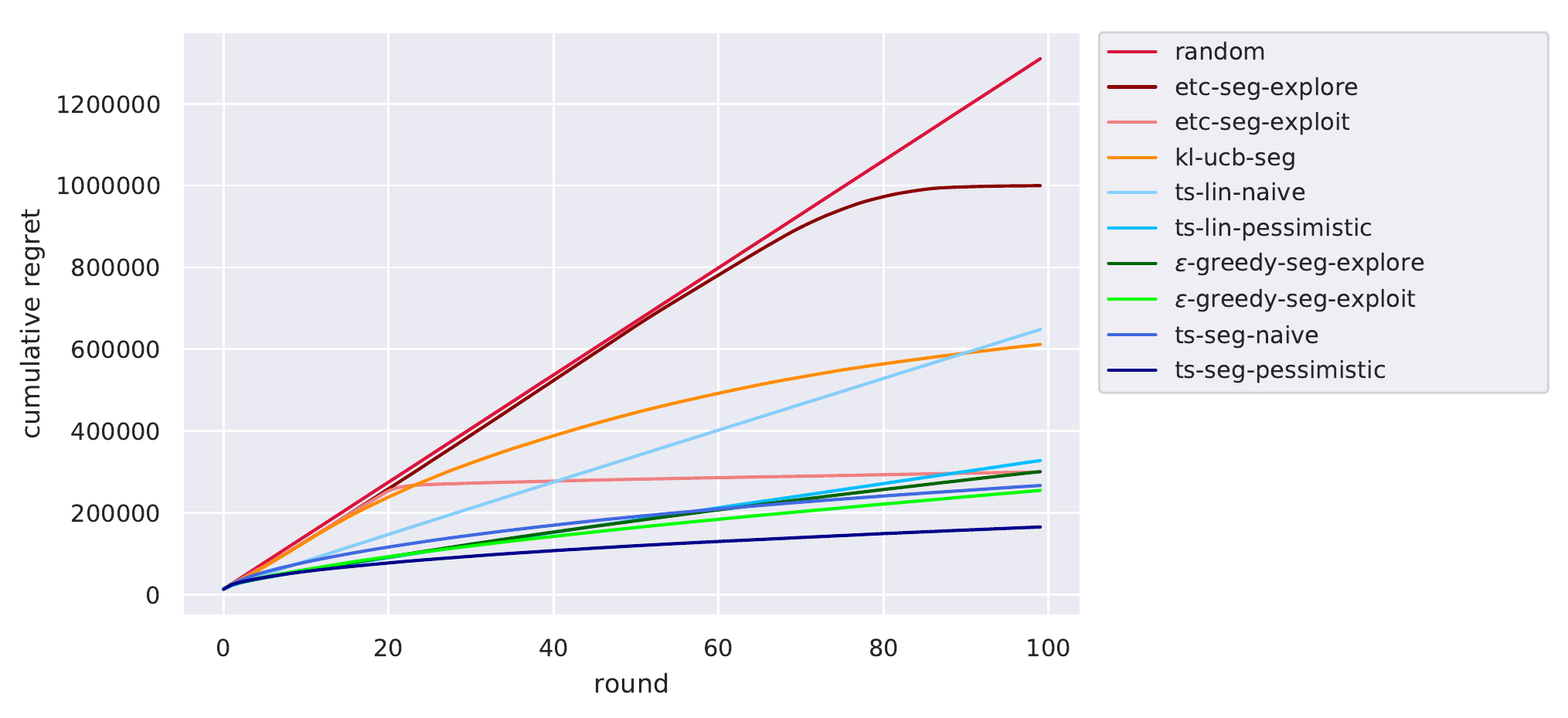}
  \vspace{-0.3cm}
  \caption{Offline evaluation of top-12 playlist recommendation: expected cumulative regrets of policies over 100 simulated rounds. The empirical gain of \textit{ts-seg-pessimistic} w.r.t. others is statistically significant at the 1\% level (p-value <0.01).}\label{fig:overall_results}
\endminipage
\end{figure*}

\subsubsection{Semi-Personalization vs Personalization}
\label{s411}

Figure \ref{fig:overall_results} provides cumulative regrets over 100 rounds for the different policies, recommending playlists via our offline environment.
Both \exploreThenCommitAlgo[-explore]{} and \exploreThenCommitAlgo[-exploit] behave as badly as \randomAlgo{} in the exploration phase, then, shortly after starting to exploit, they both reach competitive performances as illustrated by the brutal flattening of their cumulative regret curves, with \exploreThenCommitAlgo[-exploit] transitioning 50 rounds earlier. 
The later strategy also outperforms \klucbAlgo{}, which shape suggests slow learning throughout the whole experiment. 
Moreover, both \LinearTSAlgo[-pessimistic] and \LinearTSAlgo[-naive] appear to stabilize to non-flat linear cumulative regret curves after only a few rounds. Pessimistic policies are overall more effective than their naive counterparts, which is due to their lower prior display-to-stream probabilities, that are more realistic.
Overall, several semi-personalized policies eventually outclassed fully-personalized alternatives, with \TSAlgo[-pessimistic] already outperforming them all at the end of the first 25 rounds. This method manages to effectively exploit information and to quickly \textit{rank} playlists, which is an interesting result, as fully-personalized contextual models were actually the only ones able to learn the \textit{exact} display-to-stream probabilities (see generative process in Section \ref{s32}), and as both frameworks have comparable numbers of parameters ($K\times Q$ vs $K \times D$). While fully-personalized methods have been the focus of previous works on carousel recommendation \cite{gruson2019offline,mcinerney2018explore}, our experiments emphasize the empirical benefit of semi-personalization via user clustering that, assuming good underlying clusters, might appear as a suitable alternative for such large-scale real-world applications.

\subsubsection{Impact of Delayed Batch Feedback} \label{s412}
In our experiments, to be consistent with real-world constraints,  rewards are not processed on the fly but by batch, at the end of each round. We observe that, for semi-personalization, such setting tends to favor stochastic policies, such as the \textit{ts-seg} or \epsilonGreedyAlgo[] ones, w.r.t. deterministic ones such as \klucbAlgo{}. Indeed, as \klucbAlgo{} selects arms in a deterministic fashion, it always proposes the same playlists to all users of a same cluster until the round is over. On the contrary, stochastic policies propose different playlists sets within a same cluster, ensuring a wider exploration during the round, which might explain why \klucbAlgo{} underperforms in our experiments.

\subsubsection{Cascade vs No-Cascade}
\label{s413}
All policies from Figure~\ref{fig:overall_results} abide by the cascade model introduced in Section \ref{s24}. In Figure~\ref{fig:cascade_results}, we report results from follow-up experiments, aiming at measuring the empirical benefit of taking into account this cascading behaviour of users when browsing a sequence of playlists. We compared policies to alternatives that ignored the cascade display model, and thus returned a 0 reward for all unstreamed playlists at each round, whatever their rank in the carousel. Only two policy pairs are displayed in Figure~\ref{fig:cascade_results} for brevity. For both of them, the \textit{no-cascade} variant is outperformed by policies integrating our proposed cascade-based update model from Section \ref{s24}. This result validates the relevance of capturing such phenomenon for our carousel-based personalization problem.
\subsection{Online Experiments}
\label{s42}
\begin{figure*}[t]
\minipage{0.485\textwidth}
\centering
  \includegraphics[width=0.85\linewidth]{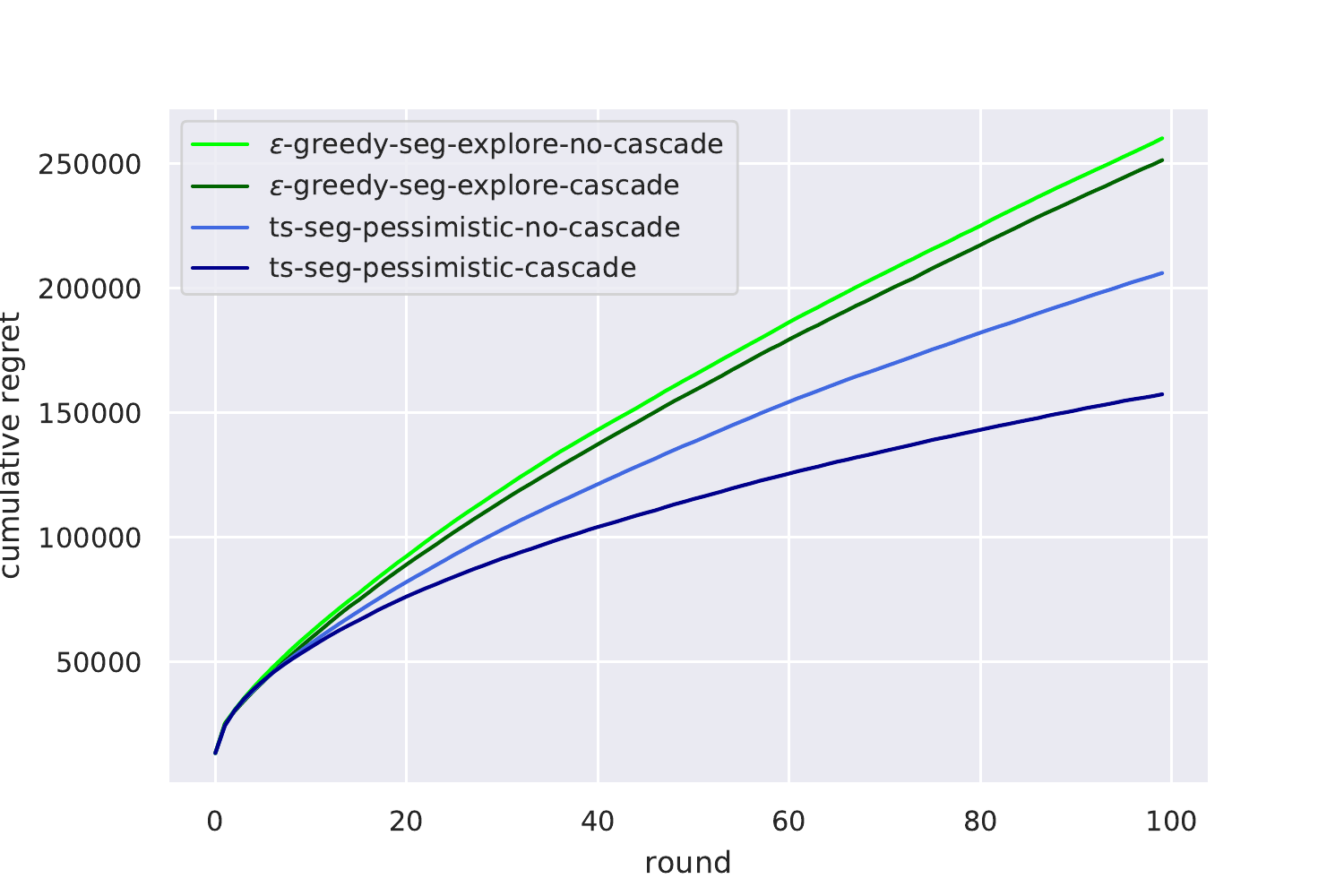}
  \vspace{-0.2cm}
  \caption{Offline evaluation: comparison of cascade vs no-cascade, over 100 simulated rounds. Differences at final round are statistically significant at the 1\% level (p-value <0.01).}\label{fig:cascade_results}
\endminipage\hfill
\minipage{0.485\textwidth}
\centering
  \includegraphics[width=0.85\linewidth]{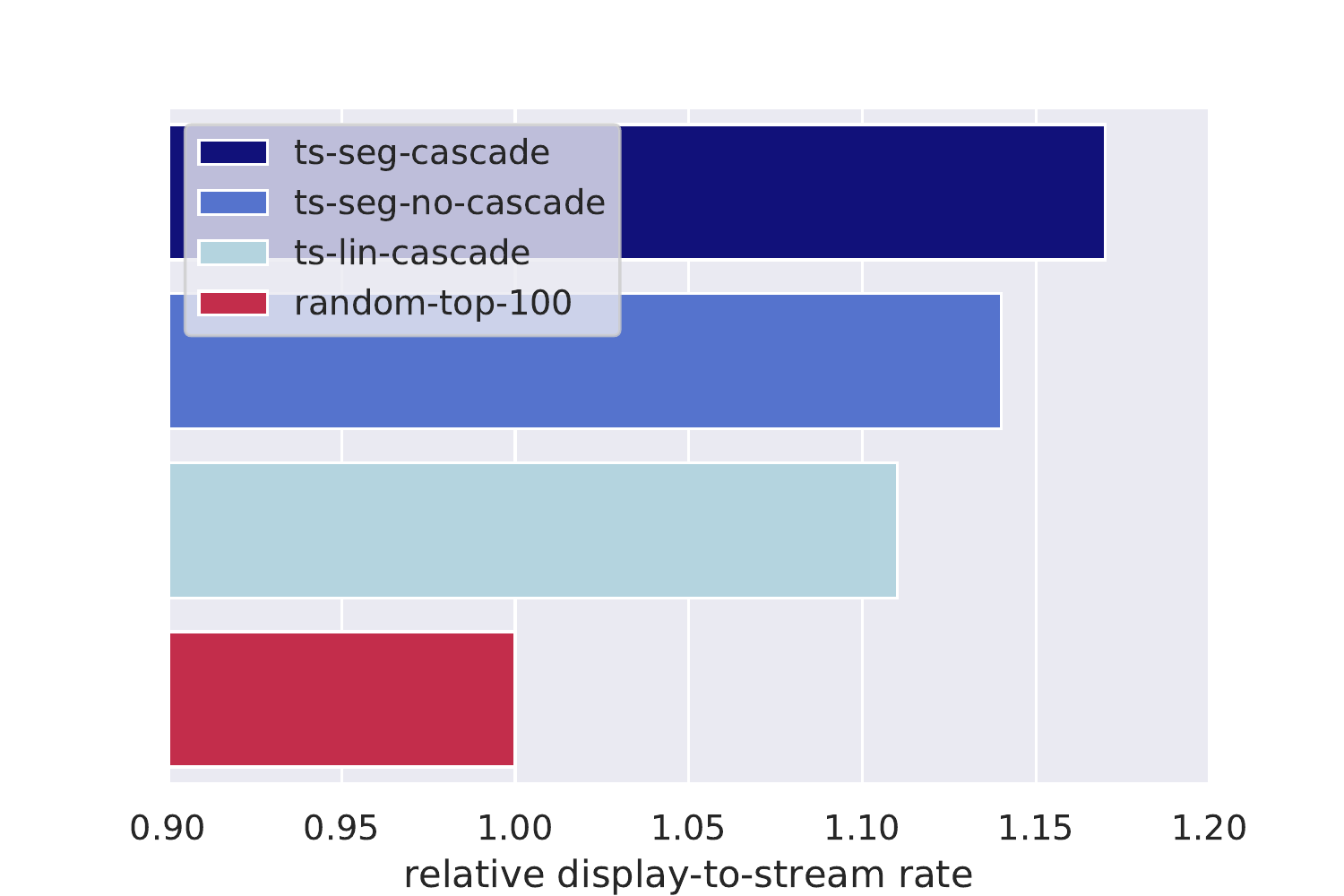}
  \vspace{-0.2cm}
  \caption{Online A/B test evaluation: relative display-to-stream gains w.r.t. \textit{random-top-100} baseline (see Section \ref{s42}). Differences are statistically significant at the 1\% level (p-value <0.01).}\label{fig:online_results}
\endminipage
\end{figure*}
An industrial-scale A/B test has been run in February 2020, to verify whether results from the simulations would hold on the actual Deezer mobile app. The 12 recommended playlists from each user's carousel were updated on a daily basis on the app. Due to industrial constraints, only a subset of policies, from (naive) Thompson Sampling, were tested in production. Also, for confidentiality reasons, we do not report the exact number of users involved in each cohort, nor the precise display-to-stream rates. Instead, results are expressed in Figure~\ref{fig:online_results} in terms of relative display-to-stream rates gains w.r.t. \textit{random-top-100}, an internal baseline that randomly recommends 12 playlists from a subset of 100, pre-selected for each cluster from internal heuristics. Results confirm the superiority of the proposed multi-armed bandit framework for personalization, notably the semi-personalized strategy, and the empirical benefit of integrating a cascade model for arms updates, although users might actually have more complex behaviours on the platform.

\section{Conclusion and Discussion}
\label{s5}

 In this paper, we modeled carousel personalization as a contextual multi-armed bandit problem with multiple plays. By addressing a challenging playlist recommendation task, we highlighted the benefits of our framework, notably the integration of the cascade model and of semi-personalization via user clustering. Along with this paper, we publicly release a large-scale dataset of user preferences for curated playlists on Deezer, and an open-source environment to recreate comparable learning problems. We believe that such release will benefit future research on carousel personalization. In particular, we assumed that the number of users and cards was fixed throughout the rounds, which is a limit, that could initiate future studies on the integration of new users or new recommendable content in swipeable carousels. Moreover, our work, as most previous efforts, also assumes that arms/cards distributions are fixed and independent, which might be unrealistic. A playlist's relative interest might depend on its neighbors in the carousel, and \textit{individually} selecting the top-$L$ playlists does not always lead to the best \textit{set} of $L$ playlists, e.g. in terms of musical diversity. Future works in this direction would definitely lead towards the improvement of carousel personalization.

\bibliographystyle{ACM-Reference-Format}
\bibliography{references}

\end{document}